\documentclass[a4paper, 10pt, conference]{ieeeconf}      
\usepackage{FG2025}
\usepackage{graphicx}
\usepackage{amsfonts}
\usepackage{amsmath}
\usepackage{float}
\usepackage{booktabs}
\usepackage{multirow}
\usepackage{hyperref}
\usepackage{makecell}

\FGfinalcopy 

\IEEEoverridecommandlockouts                              
\title{\LARGE \bf Privacy-Preserving Person Re-Identification from Temporal Sequences with Transformer and Hungarian Optimization\thanks{This is the authors' preprint version. The final authenticated version will appear in the FG 2025 proceedings published by IEEE.}}

\author{\parbox{16cm}{\centering
    {\large Raphaël Delécluse$^1$$^2$, Hazem Wannous$^1$, Laurent Guimas$^2$}\\
    {\normalsize
    $^1$ IMT Nord Europe, University of Lille, CNRS UMR 9189 - CRIStAL, F-59000 Lille, France}\\
    {\normalsize
    $^2$ Explain, F-59000 Lille, France}
}}

\usepackage{fancyhdr}
\begin{document}

\ifFGfinal
\thispagestyle{empty}
\pagestyle{empty}
\else
\author{Anonymous FG2025 submission\\ Paper ID \FGPaperID \\}
\pagestyle{plain}
\fi
\maketitle

\thispagestyle{fancy}

\begin{abstract}

Person re-identification (Re-ID) is a crucial task in surveillance and human behavior analysis, often used in public spaces such as transport hubs. Traditional RGB-based Re-ID methods raise privacy concerns and are highly sensitive to lighting variations and occlusion.  In this paper, we propose a novel Re-ID approach that leverages depth images, which inherently obscures facial and other identifiable features, making it a privacy-preserving solution. Our method addresses the association problem between multiple views of individuals by applying the Hungarian algorithm, optimizing the matching process through minimization of the global cost across the distance matrix.
We further enhance the approach by introducing temporal sequences of frames as input to a Transformer encoder architecture, which exploits both RGB and depth modalities. This architecture captures dynamic movement patterns, improving feature extraction and re-identification accuracy. Additionally, we employ batch hard triplet loss to enhance discriminative feature learning by focusing on the hardest samples. We evaluate both depth-only and RGB-D models on several top-view datasets, including TVPR2, GODPR, and BIWI RGBD-ID. Our results demonstrate that depth-only re-identification can achieve competitive performance compared to state-of-the-art methods, as measured by standard metrics such as Cumulative Matching Characteristics (CMC) and Mean Average Precision (mAP), while prioritizing privacy preservation. Code is available at: \url{https://github.com/RaphaelDel/PrivacyPreserving-ReID.git}

\end{abstract}


\section{Introduction}
Person re-identification (Re-ID) has become a fundamental task in computer vision, with widespread applications in areas such as surveillance, security, and crowd monitoring \cite{gong_person_2011,zheng_scalable_2015}. Traditional Re-ID approaches rely heavily on RGB images, which provide rich information such as color, texture, and facial features. However, the use of RGB images introduces significant privacy concerns, particularly in public spaces where individuals’ faces and other personal identifiers are captured. Furthermore, RGB-based methods are often susceptible to variations in lighting, viewpoint, and occlusion, especially in crowded environments \cite{ahmed_improved_2015}.

An alternative modality that has gained attention in recent years is depth imagery \cite{wu_robust_2017, jia_2d_2022, wu_end--end_2022}, which captures the 3D structure of a scene. Depth data, unlike RGB images, is invariant to lighting conditions and does not capture identifiable facial features, making it a more privacy-preserving option. In particular, the top-view configuration of depth sensors offers a unique advantage: it captures the full body of individuals without exposing sensitive facial information \cite{lejbolle_multimodal_2017, paolanti_person_2018}. This setup is particularly useful in applications where privacy is a priority, such as monitoring human activity in public spaces like transport hubs, shopping centers, or city squares.

In this paper, we propose a novel approach for person re-identification using only depth images, specifically in a top-view configuration. Our method leverages the geometric and anthropometric features provided by depth data to uniquely identify individuals as they move through a monitored area. To ensure accurate matching between individuals captured at different times (e.g., entering and exiting a location), we employ the Hungarian algorithm to solve the association problem. This algorithm allows us to establish optimal correspondences between incoming and outgoing individuals by minimizing the total cost of matching, effectively improving re-identification accuracy even in challenging scenarios with noises or similar body shapes.

The main contributions of this paper are threefold:

$\bullet{}$ Re-identification Using Only Depth Images: One of the key contributions of our work is the demonstration that person re-identification can be successfully achieved using only depth images. Depth data inherently preserves privacy by omitting facial or other identifiable features, making it ideal for use in public spaces. Despite lack of RGB information, we show that depth-based models can achieve competitive re-identification performance, particularly when combined with appropriate techniques for feature extraction and matching.

$\bullet{}$ Utilization of Temporal Sequences for Enhanced Re-Identification: Rather than using individual frames for re-identification, our approach processes sequences of frames, which allows us to capture the temporal dynamics and movement patterns of individuals. This use of sequences provides more robust feature representations, as body postures and walking patterns offer additional information that can distinguish individuals more effectively than a single snapshot. Both our depth-only model and the RGB-D model benefit from this sequential processing, significantly improving the accuracy of the re-identification process. This sequential approach enhances performance, particularly when individuals exhibit similar appearances across frames or when the scene contains occlusions or noise.

$\bullet{}$ Optimization Using the Hungarian Algorithm: By integrating the Hungarian algorithm, we optimize the association process across multiple views, ensuring robust and precise re-identification even in complex scenarios. While our RGB-D model is included as a point of comparison, the primary focus of this work is to demonstrate the feasibility and utility of depth-only Re-ID in privacy-sensitive applications.

To demonstrate the effectiveness of our method, we evaluate it on several depth-based datasets, comparing its performance to traditional RGB-D approaches. Our results show that depth-only re-identification can achieve competitive accuracy, especially in a top-view configuration when combined with the Hungarian algorithm for optimizing the association of individuals between different time frames.

In summary, this work highlights the potential of using depth images for privacy-preserving person re-identification. By integrating the Hungarian algorithm to solve the association problem, and leveraging the power of temporal sequences, our approach offers a practical solution for accumulating valuable data in applications such as transport network optimization, where preserving the privacy of individuals is paramount.

\section{Related Works}
\label{sec:related_works}

Person re-identification has emerged as a pivotal task in video surveillance, security, and human-computer interaction, owing to the increasing need for systems capable of tracking individuals across different camera views \cite{ye_deep_2022, leng_survey_2020}. The traditional approaches to Re-ID have primarily focused on RGB-based data, but recent developments have extended the problem to other modalities such as depth \cite{wu_robust_2017} and infrared (IR) \cite{wu_rgb-infrared_2017} data, as well as various camera perspectives like top-view, which introduce new challenges and opportunities for enhancing performance.

\subsection{Person Re-Identification from RGB Data}
RGB-based person Re-ID methods have been the cornerstone of research in this area for many years \cite{rao_hierarchical_2024}. These approaches utilize deep learning techniques to extract discriminative features from RGB images, aiming to learn invariant representations that can generalize across different viewpoints, lighting conditions, and occlusions \cite{ye_deep_2022}. Common architectures include Convolutional Neural Networks (CNNs) such as ResNet, which have been shown to be effective for this task. Variants such as part-based models \cite{ferrari_beyond_2018} and attention mechanisms \cite{chen_self-critical_2019} have been proposed to improve feature extraction by focusing on important body parts or regions. However, despite these advancements, RGB-based Re-ID still struggles with occlusions and changes in appearance, particularly in crowded scenes or when capturing individuals from non-traditional angles, such as top-down perspectives.

\subsection{Re-Identification from Depth Data}
The limitations of RGB-based methods have led to the exploration of alternative modalities, with depth data being one of the most promising. Depth sensors provide a 3D structure of the scene, capturing the geometric properties of individuals that are invariant to changes in lighting and clothing, offering a complementary cue to RGB images \cite{busto_open_2017}. Depth-based methods can leverage the skeletal and contour information provided by depth sensors to improve recognition in scenarios where RGB data might be unreliable or insufficient \cite{rao_hierarchical_2024, rao_simmc_2022, rao_transg_2023}. For example, Si et al. \cite{ferrari_skeleton-based_2018} proposed using skeleton data derived from depth maps to represent individuals, thus mitigating occlusion issues.

Despite the potential of depth-based Re-ID, there has been relatively less research in this domain compared to RGB-based methods, and large-scale depth datasets for Re-ID remain scarce. Nevertheless, depth-based approaches are particularly useful in overhead top-view setups, where body shape and posture can be key distinguishing features.

\subsection{Top-View Person Re-Identification}
Top-view Re-ID presents unique challenges as the visibility of distinguishing features, such as the face or torso, is significantly reduced. This perspective is increasingly relevant in surveillance systems, where overhead cameras are used to minimize occlusions and protect privacy by avoiding facial recognition \cite{liciotti_person_2017}. The task becomes one of identifying individuals based on body shape, movement, and appearance cues from above. 

Methods designed for top-view Re-ID need to adapt to these constraints by focusing on features that are visible from this angle. Depth data proves particularly valuable in this context, as it provides additional structural information that RGB alone cannot capture \cite{paolanti_sesame_2022}. The TVPR2 dataset, introduced by \cite{martini_open-world_2020}, is specifically designed for evaluating top-view Re-ID models using both RGB and depth information, making it a benchmark for research in this area.

\subsection{Multi-Modal Person Re-ID:}
One of the most promising approaches to Re-ID in recent years has been the fusion of multiple modalities \cite{wu_rgb-infrared_2017, luna_people_2021}, particularly the combination of RGB and depth data \cite{hafner_cross-modal_2022, wu_end--end_2022}. By combining the strengths of each modality—RGB providing texture and color information, and depth offering geometric invariance— previous works have achieved improved performance in Re-ID tasks. 
Pala et al. \cite{pala_multimodal_2016} extracted global features using adaptive average pooling and embedded them into local part features by computing relational similarities. This method effectively fuses RGB and depth data, leveraging the complementary strengths of texture and geometry to enhance person re-identification performance across varied viewpoints.

Multi-modal methods typically concatenate or fuse features from RGB and depth streams, using architectures such as two-stream CNNs with fusion layers \cite{ren_multi-modal_2017} or attention-based models \cite{uddin_depth_2020, lejbolle_attention_2018, lejbolle_person_2020, mukhtar_cmot_2024} to focus on the most relevant parts of each modality. However, creating efficient models that can seamlessly integrate multi-modal data remains a challenge, especially when the data modalities exhibit differing levels of noise or reliability.

\subsection{Datasets for Depth and Top-View Person Re-ID}
The availability of robust datasets is essential for advancing Re-ID research. While numerous datasets exist for RGB-based Re-ID, such as Market-1501 \cite{zheng_scalable_2015}, DukeMTMC-ReID \cite{ristani_performance_2016,zheng_unlabeled_2017}, and CUHK03 \cite{li_deepreid_2014}, datasets focused on depth and top-view Re-ID are relatively limited. 

The TVPR2 Dataset \cite{martini_open-world_2020}, is one of the most extensive datasets for top-view Re-ID, containing over 1000 unique individuals captured using RGB-D sensors. This dataset provides a crucial resource for evaluating multi-modal Re-ID methods in top-view settings, where the challenges differ significantly from traditional side-view scenarios.

Another important dataset is the GODPR Dataset (Geintra Overhead Depth People Re-identification) \cite{fuentes-jimenez_depth_2020}, which includes both infrared and depth data captured from an overhead perspective. Although initially designed for detection tasks, GODPR has become an important resource for developing and testing Re-ID models that rely on depth and IR modalities. This dataset is particularly challenging due to the use of IR data, which requires models to adapt to a different form of visual input.

The BIWI RGBD-ID Dataset \cite{munaro_one-shot_2014}, another significant dataset in the field, focuses on long-term person re-identification using both RGB and depth data. It includes synchronized RGB images, depth images, and skeletal data, providing a diverse set of features for multi-modal Re-ID. The BIWI dataset has been widely used to test models that rely on body cues rather than facial features, as it primarily captures individuals from a frontal perspective, making it complementary to the TVPR2 and GODPR datasets, which use top-view configurations.

The field of person Re-ID is rapidly evolving, with increasing interest in multi-modal and top-view approaches that address the limitations of traditional RGB-based methods. The fusion of depth and RGB data, combined with advanced neural architectures such as attention mechanisms and two-stream networks, holds great promise for improving Re-ID performance in challenging real-world scenarios.

\section{Methodology}
\label{sec:methodology}

\subsection{Challenge framing}
    
    Our methodology operates under a hypothesis designed to closely mimic real-world surveillance and monitoring applications. Specifically, we assume that each individual captured by the system passes under the camera twice—once while entering a monitored area and once while exiting. This assumption is grounded in typical scenarios such as entrance-exit tracking, where the same individuals appear in different temporal sequences. Consequently, each unique individual is associated with two distinct sequences of images: one from the time of entry and the other from the time of exit. The task of person re-identification (Re-ID) in this context is to establish correspondence between these entry and exit sequences.
    
    Given that the datasets provides explicit labels for each individual, allowing us to know which sequences belong to the same person, the challenge becomes one of associating the correct sequences from both directions (i.e., matching incoming individuals to their outgoing counterparts). This setup mirrors practical situations where individuals are observed at multiple points in time and space, with the goal of re-identifying them across these different observations.
    
    To facilitate the re-identification process, our model takes as input a sequence of Regions of Interest (RoIs) extracted from depth images. These RoIs are obtained through a robust blob detection algorithm, which is both simple and effective in isolating individuals from the background based on depth information. By focusing on depth-based RoIs, we exploit the invariant properties of depth images, such as resistance to changes in lighting or clothing, ensuring that our model generates reliable hypotheses for matching sequences. This forms the foundation of our Re-ID system, where the primary goal is to correlate the in-and-out image sequences for each unique individual based on the extracted RoIs from depth data.
    
    \subsection{Transformer based approach}
    
    \begin{figure}
        \centering
        \includegraphics[width=0.5\textwidth]{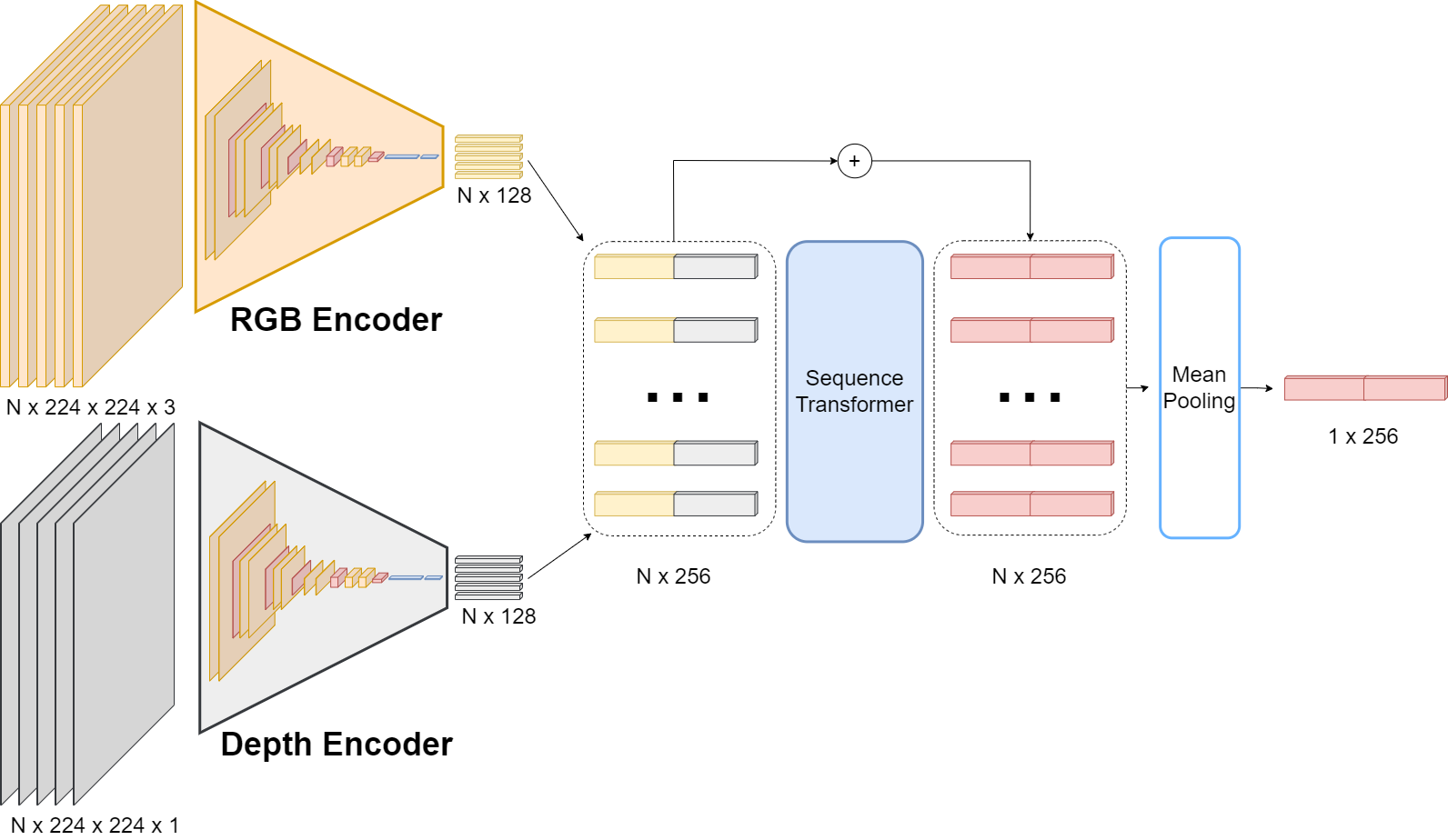}
        \caption{\textbf{RGB-D Re-Identification Model Architecture} —- The model utilizes two separate encoders: an RGB encoder and a depth encoder. Both encoders output embeddings of size \textit{N×128}, which are then concatenated to form a joint representation of size \textit{N×256}. The sequence of concatenated embeddings is passed through a transformer to model temporal dependencies. A final mean pooling operation is applied to generate a single embedding of size \textit{1×256}, which is used for the re-identification task.}
        \label{fig:RGBD_model}
        \vspace{-0.5cm}
    \end{figure}
    
Our re-identification approach leverages a transformer-based architecture to project sequences of Regions of Interest (RoIs) into a latent space, where similar individuals are mapped close to each other, and distinct individuals are mapped further apart. The goal is to utilize both spatial and temporal information from the sequence of RoIs to improve the accuracy of re-identification.
    
We propose two versions of our approach. The first incorporates both RGB and depth information, while the second considers only depth data, prioritizing privacy preservation. In section \ref{sec:experiments}, we will compare the performance of these two methods and evaluate the contribution of each modality to the overall performance.

Our model architecture, as shown in Figure \ref{fig:RGBD_model}, is composed of two primary components:\\
    
$\bullet{}$ \textbf{RoI Feature Encoding}: In the first stage, we encode the information contained within each RoI, which has been pre-extracted from RGBD images, into an embedding of size \( D \). For the RGB stream, we used a ResNet-50 \cite{he_deep_2016} architecture pre-trained on ImageNet \cite{deng_imagenet_2009}. This model is well-suited for extracting high-level visual features from RGB images. For the depth stream, we utilize a smaller variation of ResNet \cite{he_deep_2016}, which is not pretrained due to the limited availability of large-scale depth image datasets. The depth encoder is designed with a single input filter to process the depth information. The use of bottleneck layers and residual connections in both encoders significantly reduces the number of parameters and accelerates the convergence of the model by improving the gradient flow \cite{szegedy_inception-v4_2017}.

$\bullet{}$ \textbf{Sequence Encoding via Transformers}: In the second stage, we take the embeddings produced from both the RGB and depth streams and use them as inputs to a transformer encoder \cite{vaswani_attention_2017} to capture the temporal dynamics of the sequence. For each frame, we obtain two embeddings—one for the RGB image and one for the corresponding depth image—yielding a sequence of \( 2N \) embeddings, where \( N \) is the number of frames. These embeddings are concatenated into \( N \) embeddings of size \( 2D \), representing the combined RGB and depth information for each frame.

While our approach is multi-modal, fusing both RGB and depth data, we deliberately adopt a simple concatenation strategy for fusing these modalities rather than employing cross-attention mechanisms. This decision is motivated by the role of the RGB-D model in this study, which serves primarily as a point of comparison for the depth-only model. Given this context, concatenation offers a computationally efficient and effective solution for integrating the two modalities.

The concatenation strategy allows the transformer encoder to model both modalities as a unified sequence, leveraging self-attention to extract temporal dependencies across the RGB and depth information simultaneously. By treating both modalities equally within the temporal sequence, the self-attention mechanism captures the significance of each frame without requiring the additional complexity of cross-attention. This design ensures a streamlined architecture that effectively benefits from multi-modal information while remaining computationally efficient.

The self-attention mechanism, as introduced by Vaswani et al. \cite{vaswani_attention_2017}, allows the model to capture long-range dependencies across the entire sequence, effectively weighing the significance of each frame in relation to the others. 

Finally, we apply a mean pooling operation across the sequence embeddings to improve the model’s resilience to noise, such as frame occlusions or sensor irregularities, ensuring that the final embedding generalizes well to unseen data.

This transformer-based architecture, which processes RGB and depth jointly without the added complexity of cross-attention, enables the model to learn both spatial and temporal patterns efficiently, resulting in a robust and accurate re-identification system, even in scenarios with challenging or incomplete data inputs.

\subsection{Batch Hard Triplet Losses}

Triplet loss, as introduced in works such as \cite{schroff_facenet_2015}, has become a standard approach in metric learning, particularly for person re-identification tasks. This loss function encourages the network to learn embeddings that ensure positive samples (same identity as the anchor) are closer to the anchor than negative samples (different identity). Several notable works \cite{hermans_defense_2017, zhao_deep_2020} have demonstrated the effectiveness of triplet loss for tasks where the goal is to separate instances based on learned features. However, despite its success, the classical implementation of triplet loss suffers from several limitations, particularly in large-scale re-identification scenarios. However, the classical triplet loss struggles with large-scale scenarios due to inefficient sampling of informative triplets, which slows convergence and limits generalization.

Alternative strategies, such as the \textbf{Batch Hard Triplet Loss} \cite{hermans_defense_2017}, a more efficient and effective approach that focuses on the most challenging triplets within each batch. Instead of relying on pre-selected or random triplet pairs, this method selects the hardest positive (i.e., the sample most similar to the anchor from the same identity) and hardest negative (i.e., the sample most dissimilar to the anchor from a different identity) within each mini-batch, ensuring that the network is trained with the most informative and difficult examples. This strategy helps the model focus on challenging cases, thus enhancing its robustness and generalization performance.

Formally, in our approach, \( f_{\theta}(x): \mathbb{R}^F \to \mathbb{R}^D \) represents a neural network that maps an input image \( x \) from the input space \( \mathbb{R}^F \) (where \( F \) is the feature dimension) to an embedding space \( \mathbb{R}^D \) (where \( D \) is the embedding dimension). The goal of \( f_{\theta} \) is to learn embeddings that place semantically similar points (such as images of the same person) close together and dissimilar points farther apart in this embedding space. We form batches by sampling \( P \) classes (person identities) and \( K \) images per class, resulting in a batch of \( PK \) images. For each anchor sample \( x_a^i \), we compute the hardest positive and hardest negative in the batch, defining the loss as follows in equation \ref{eq:hard_loss}:

\vspace{-0.5cm}
\begin{equation}\label{eq:hard_loss}
  \begin{split}
    L_{BH}(\theta; X) = \sum_{i=1}^{P} \sum_{a=1}^{K} \biggl\{
    &m + \max_{p=1,...,K} D(f_{\theta}(x_a^i), f_{\theta}(x_p^i)) \\
    & - \min_{\substack{j=1,...,P \\ n=1,...,K \\ j \neq i}} D(f_{\theta}(x_a^i), f_{\theta}(x_n^j)) \biggr\}
  \end{split}
\end{equation}

Here, \( D \) represents the Euclidean distance between embeddings, \( x_a^i \) is the anchor sample, \( x_p^i \) is the hardest positive, and \( x_n^j \) is the hardest negative, with \( m \) controlling the margin between positive and negative pairs.

By focusing on the most difficult examples in the batch, the Batch Hard Triplet Loss ensures faster convergence and improves the model's ability to generalize to difficult re-identification scenarios. This is particularly beneficial in cases where individuals may exhibit similar body shapes or movement patterns, as is common in top-view re-identification settings.
    
To further assist with the convergence of our multi-modal architecture, we utilize a combination of triplet losses:
    
    \begin{itemize}
        \item \textbf{Transformer Loss}: Applied to the final output of the model, capturing the temporal and multi-modal information from the transformer.
        \item \textbf{Depth Loss}: Applied to the output of the depth encoder to ensure the network learns discriminative depth features.
        \item \textbf{RGB Loss}: Applied to the output of the RGB encoder when the model is operating in RGB-D mode, ensuring the network learns discriminative RGB features.
    \end{itemize}
    
    In the RGB-D configuration, the final loss is the sum of these three components, as it can be seen in equation \ref{eq:triplet_triplet}:
    \begin{equation}\label{eq:triplet_triplet}
    L_{\text{final}} = L_{\text{Transformer}} + L_{\text{Depth}} + L_{\text{RGB}}
    \end{equation}
    
    This loss formulation encourages the model to learn complementary features from the RGB and depth modalities, while the transformer loss focuses on sequence-level information. By incorporating these multiple loss terms, we prevent either the RGB or depth modality from dominating the feature extraction process, ensuring a balanced contribution from both modalities in the re-identification task.
    
\subsection{Hungarian Optimization}\label{sec:assignment_problem}

    \begin{figure}
    \centering
    \includegraphics[width=0.4\textwidth]{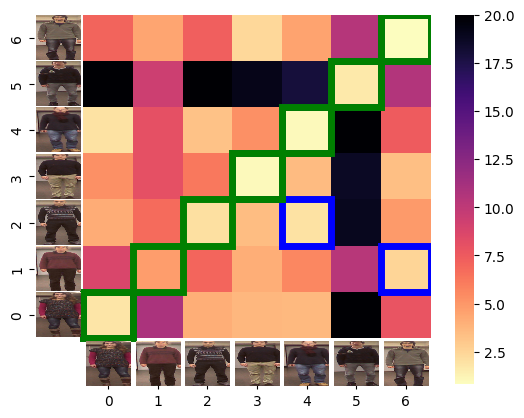}
    \caption{Illustration of the Hungarian algorithm \cite{kuhn_Hungarian_1955} solving the assignment problem in context of a cross-dataset evaluations in BIWI RGBD-ID dataset. The heatmap represents the distance matrix where each cell denotes the pairwise distance between query and gallery embeddings. The blue squares indicate the initial matches based on minimum distances, while the green squares highlight corrected matches made by the Hungarian algorithm.}
    \label{fig:Hungarian_improvement}
    \vspace{-0.5cm}
    \end{figure}

In our person re-identification pipeline, once the embeddings have been extracted from the test sequences, we face the problem of correctly matching the query set (incoming individuals) to the gallery set (outgoing individuals). This matching task is formulated as an assignment problem, where each query embedding must be paired with its correct counterpart in the gallery based on a distance matrix. The Hungarian algorithm \cite{kuhn_Hungarian_1955} is employed to solve this assignment problem efficiently.

\paragraph{Application of the Algorithm} The Hungarian algorithm is applied in the following steps:

$\bullet{}$ Distance Matrix Construction: A distance matrix  is constructed, where  represents the Euclidean distance between the embedding of query  and gallery.

$\bullet{}$ Input to the Algorithm: The algorithm processes the distance matrix to identify the optimal one-to-one assignments that minimize the total cost (sum of selected distances).

$\bullet{}$ Output: The result is a set of matched pairs, where each query embedding is uniquely assigned to a gallery embedding, ensuring no conflicts.

\paragraph{Advantages of the Algorithm} The Hungarian algorithm optimizes the global assignment problem by considering all possible matches simultaneously, unlike nearest-neighbor approaches that operate locally. This ensures:

$\bullet{}$ Conflict Avoidance: Prevents multiple queries from being assigned to the same gallery embedding.

$\bullet{}$ Global Optimization: Accounts for the full context of the dataset, improving robustness in scenarios with noise or similar appearances.

\paragraph{Preprocessing and Assumptions} Embeddings are normalized before constructing the distance matrix to ensure consistency in feature space. The algorithm assumes a one-to-one correspondence between the query and gallery sets, simplifying the assignment process and guaranteeing unique matches for every query.

\paragraph{Impact on Re-Identification Performance} The Hungarian algorithm significantly enhances re-identification performance by optimizing the matching process. For example, in our cross-dataset evaluation on BIWI RGBD-ID, precision increased by 8\% for RGBD and 25\% for depth compared to nearest-neighbor matching. Figure \ref{fig:Hungarian_improvement} illustrates how the algorithm resolves conflicts, improving the overall accuracy of the system.
By addressing the global assignment problem holistically, the Hungarian algorithm ensures optimal and precise matching across all datasets and conditions, making it indispensable for our privacy-preserving, depth-only re-identification system.

\section{Experiments}
\label{sec:experiments}

\subsection{Datasets}
 
\subsubsection{\textbf{TVPR2 Dataset}}
        \begin{figure}
        \centering
        \includegraphics[width=0.42\textwidth]{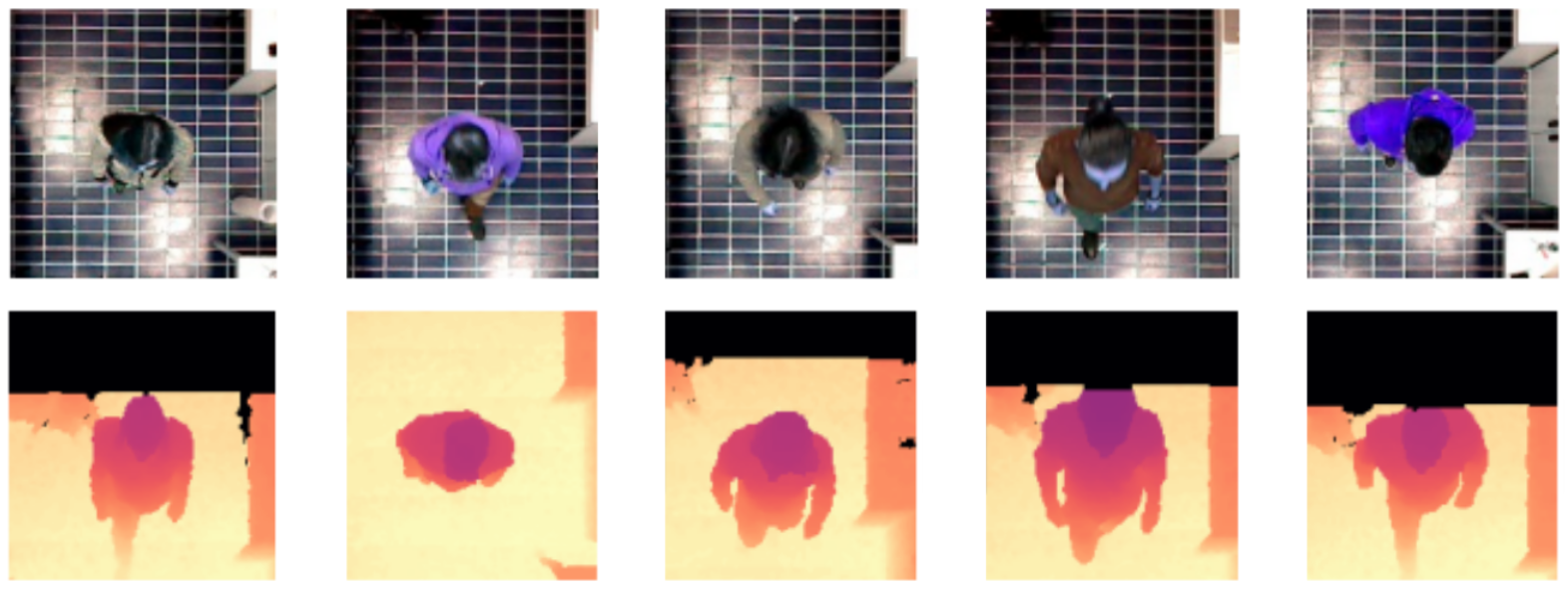}
        \caption{Illustration of the different modalities of the TVPR2 dataset: RGB and depth images from a top-view perspective.} \label{fig:tvpr_dataset}
        \vspace{-0.5cm}
        \end{figure}

The TVPR2 \cite{martini_open-world_2020} (Top-View Person Re-identification 2) dataset was created for the specific purpose of evaluating person Re-ID models in top-view surveillance scenarios. It contains 235 video sequences, featuring 1027 unique individuals captured in a controlled indoor environment. Both RGB and depth information are recorded, as shown in figure \ref{fig:tvpr_dataset}, allowing for multi-modal Re-ID approaches. The dataset simulates real-world surveillance settings, with individuals walking beneath a ceiling-mounted camera. The top-view configuration ensures privacy preservation by avoiding facial details, while capturing full-body data for robust Re-ID. In addition, TVPR2 includes various sessions recorded over eight days, featuring variations in group sizes, natural lighting conditions, and demographic backgrounds. These factors make TVPR2 ideal for testing Re-ID methods that must operate in variable conditions with diverse human appearances.
    
\subsubsection{\textbf{GODPR Dataset}}
The GODPR (Geintra Overhead Depth People Re-identification) dataset, introduced by Fuentes-Jimenez et al. \cite{fuentes-jimenez_depth_2020}, focuses on person Re-ID using overhead, top-view RGB-D sensors in diverse real-world conditions. It comprises 136 sequences captured with two types of depth sensors: Kinect V2 (time-of-flight) and Intel RealSense D435 (active stereo). The dataset is divided into three subsets, each representing different environmental conditions, sensor heights, and participant characteristics, such as the presence of face masks. 
    
\subsubsection{\textbf{BIWI RGBD-ID Dataset}}
    
        \begin{figure}
        \centering
        \includegraphics[width=0.4\textwidth]{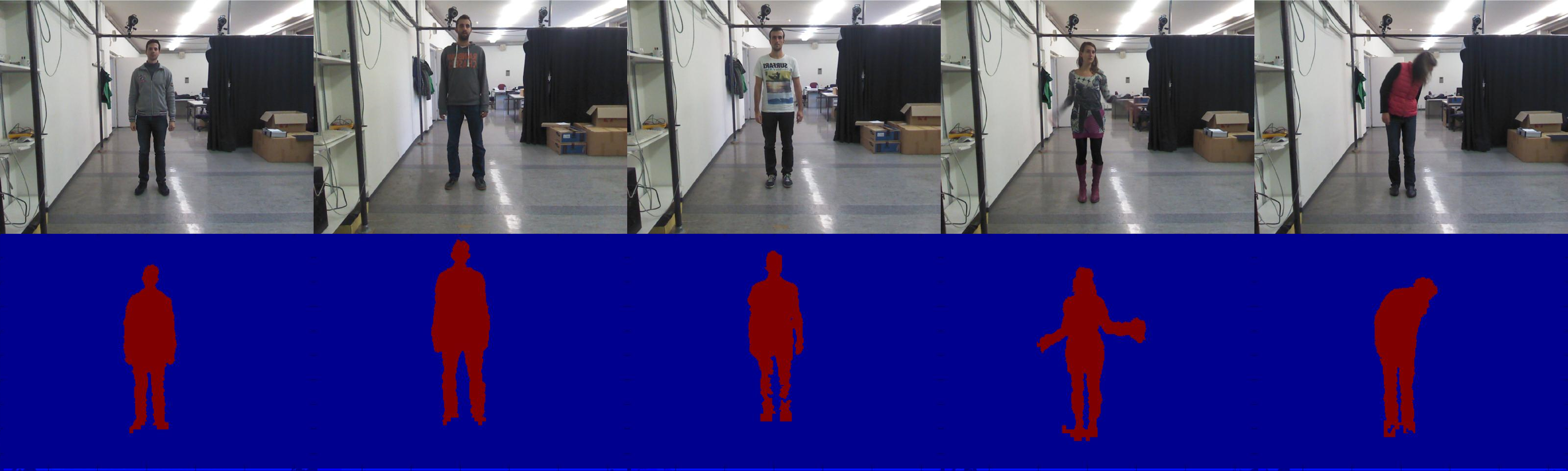}
        \caption{Illustration of the different modalities of the BIWI RGBD-ID dataset: RGB and depth images.} \label{fig:biwi_dataset}
        \vspace{-0.5cm}
        \end{figure}

The BIWI RGBD-ID dataset \cite{munaro_one-shot_2014} is designed for long-term person re-identification using synchronized RGB and depth images, collected with a Microsoft Kinect sensor, as is can be seen in figure \ref{fig:biwi_dataset}. This dataset captures 78 individuals across 106 sequences, split between training and testing sets. It includes both static and dynamic movements, such as standing still or walking, providing a variety of postures and orientations. Unlike TVPR2, which is top-view, the BIWI dataset focuses on front-facing recordings, emphasizing body cues rather than facial features. This makes it well-suited for evaluating Re-ID models that rely on skeletal and anthropometric features in constrained, frontal-view scenarios. BIWI’s inclusion of synchronized RGB, depth, skeletal data, and ground plane coordinates offers a comprehensive benchmark for testing multi-modal and depth-based Re-ID approaches.
Table \ref{tab:datasets} summarizes these datasets. 

    \begin{table}[H]
        \caption{Summary of the datasets}\label{tab:datasets}
        \tabcolsep=0.11cm
        \resizebox{\columnwidth}{!}{%
        \begin{tabular}{llll}
        Dataset &  TVPR2 & GODPR & BIWI RGBD-ID\\
        \hline
        \\[-0.75em]
        Sensor & Asus Xtion Pro Live & \makecell[l]{Kinect 2 \&\\ RealSense D435} & Kinect 1\\
        \hline
        \\[-0.75em]
        Data Modality & RGB-D & IR-D & RGB-D\\
        \hline
        \\[-0.75em]
        Total individuals & 1027 & 68 & 78 \\
        \hline
        \\[-0.75em]
        Type & Structured light & \makecell[l]{Time of flight \&\\ Active stereo} & Time of flight\\
        \hline
        \end{tabular}%
        }
        \vspace{-0.3cm}
    \end{table}



\subsection{Evaluations}

    \subsubsection{\textbf{Evaluation protocol}}
        \hfill\\
        We evaluated our person re-identification method using standard metrics such as Cumulative Matching Characteristics (CMC), Mean Average Precision (mAP), and Precision\cite{paolanti_deep_2020, luna_people_2021, mukhtar_cmot_2024}.
        
        \paragraph{Cumulative Matching Characteristics (CMC)} CMC-k (Rank-k) represents the probability that the correct match appears within the top-k ranked results. Specifically, CMC-1 indicates the likelihood that the correct identity is the highest-ranked result.
        
        \paragraph{Mean Average Precision (mAP)} mAP averages the precision at different recall levels across all queries. It evaluates how well the model ranks relevant images higher in retrieval tasks.
        
        \paragraph{Precision and Distance Matrix} Precision measures the effectiveness of the Hungarian algorithm in solving the assignment problem. We compute a distance matrix from the extracted embeddings, as it is showed in figure \ref{fig:Hungarian_improvement}, where CMC-1 corresponds to correctly matched diagonal elements (minimum distances). The Hungarian algorithm is then used to optimize matching across the entire test set.

\hfill\\[-1em]
\subsubsection{Comparison between depth and RGB-D approach}

    \begin{table}
        \centering
        \caption{Results obtained by our models on TVPR2, GODPR and BIWI RGBD-ID (in \%, higher is better)}\label{tab:results}
        \tabcolsep=0.11cm
        \resizebox{\columnwidth}{!}{%
        \begin{tabular}{|l|l|l|l|l|l|l|}
        \hline
        Dataset &  Modality & \textbf{Precision} & \textbf{Rank-1} & Rank-5 & Rank-10 & \textbf{mAP}\\
        \hline
        \hline
        TVPR2 & RGB-D & 100.0 & 99.5 & 100.0 & 100.0 & 99.9\\
              & Depth & 94.6 & 88.4 & 98.1 & 99.5 & 96.6\\
        \hline
        GODPR & IR-D & 100.0 & 100.0 & 100.0 & 100.0 & 98.4\\
              & Depth & 100.0 & 78.6 & 100.0 & 100.0 & 68.0\\
        \hline
        BIWI RGBD-ID & RGB-D & 100.0 & 96.4 & 100.0 & 100.0 & 97.4\\
                     & Depth & 100.0 & 60.7 & 96.4 & 100.0 & 56.6\\
        \hline
        \end{tabular}}%
        \vspace{-0.5cm}
    \end{table}

    We compare here the performance of our approach using depth-only and RGB-D modalities across the three datasets discussed earlier. Obtained results, shown in Table \ref{tab:results}, highlight the impact of incorporating RGB data in addition to depth information.
    
    We observe that the re-identification performance consistently improves when using RGB-D or IR-D modalities compared to depth alone. For the TVPR2 dataset, the Rank-1 accuracy increased significantly from 88.4\% with depth-only to 99.5\% with RGB-D. Similarly, for the GODPR dataset, the Rank-1 accuracy improved from 78.6\% to 100.0\%, and for the BIWI RGBD-ID dataset, it increased from 60.7\% to 96.4\%.
    
    Although the improvements may seem substantial, the performance gap can be mitigated using the Hungarian algorithm, which enhances the re-identification accuracy by solving the assignment problem. For the GODPR and BIWI RGBD-ID datasets, this algorithm enabled perfect re-identification, achieving 100.0\% precision. In the case of the TVPR2 dataset, the improvement, while less pronounced, still raised the precision from 88.4\% to 94.6\%. The Hungarian algorithm's effectiveness is somewhat diminished for TVPR2 due to the larger size and complexity of the test set, as discussed in Section \ref{sec:ablation_study}.

    While depth-only models prioritize privacy by omitting texture and color information, they face inherent limitations in differentiating individuals with similar body shapes or movement patterns. These constraints become particularly evident in datasets like BIWI, where the RGBD model achieves a rank-1 accuracy of 96.4\%, compared to 60.4\% for the depth-only model. The BIWI dataset differs from others as it is not top-view, highlighting the challenges of applying a depth model trained on top-view perspectives, such as TVPR2, to different angles. Furthermore, depth-only models struggle to capture fine-grained details, such as clothing texture or subtle appearance cues, and are more susceptible to noise in depth data, which can degrade feature extraction and matching accuracy.

    Despite these challenges, the proposed method demonstrates that depth-only re-identification can achieve competitive performance in privacy-sensitive scenarios, especially when combined with robust optimization techniques such as the Hungarian algorithm. This is particularly evident in the BIWI dataset, where the Hungarian algorithm ensures a precision of 100.0\% for both the RGBD and depth-only models, despite the significant gap in rank-1 accuracy. By leveraging global optimization, the algorithm mitigates the challenges posed by the dataset’s different viewing angles, ensuring robust matching even in suboptimal conditions.

\hfill\\[-1em]
\subsubsection{Comparison with State of the Art}

    \begin{table}
        \centering
        \caption{Comparative results of our RGBD approach and SOTA on TVPR2, BIWI RGBD-ID and GODPR (in \%, higher is better)}\label{tab:SOTA_comparaison}
        \tabcolsep=0.11cm
        \resizebox{\columnwidth}{!}{%
        \begin{tabular}{|l|l|l|l|l|l|l|}
        \hline
        Dataset &  Method & \textbf{Precision} & \textbf{Rank-1} & Rank-5 & Rank-10 & \textbf{mAP} \\
        \hline
        \hline
        TVPR2 & VRAI-Net3\cite{paolanti_deep_2020} & 74.4 & 74.4 & - & - & 77.9 \\
        & MAT\cite{lejbolle_attention_2018} & 91.1 & 91.1 & 93.7 & 94.1 & 78.8 \\
        & SLATT\cite{lejbolle_person_2020} & 90.6 & 90.6 & 92.1 & 94.3 & 75.1 \\
        & TL-DCNN\cite{martini_open-world_2020} & 94.0 & 94.0 & 96.6 & 97.1 & 93.3 \\
        & SeSAME\cite{paolanti_sesame_2022} & 98.8 & 98.8 & 99.4 & 99.7 & 81.6 \\
        & CMOT\cite{mukhtar_cmot_2024} & 99.1 & 99.1 & 99.1 & 99.9 & 94.2 \\
        & Ours & \textbf{100.0} & \textbf{99.5} & \textbf{100.0} & \textbf{100.0} & \textbf{99.9} \\
        \hline
        GODPR & Luna \& al.\cite{luna_people_2021} & 95.5 & 95.5 & - & - & - \\
        & Ours & \textbf{100.0} & \textbf{100.0} & \textbf{100.0} & \textbf{100.0} & \textbf{98.4} \\
        \hline
        BIWI RGBD-ID
        & HRN\cite{wu_end--end_2022} & 47.1 & 47.1 & 73.5 & 78.1 & 44.6 \\
        & Distillation\cite{hafner_cross-modal_2022} & 40.4 & 40.4 & 77.1 & 91.0 & 41.3 \\
        & SimMC\cite{rao_simmc_2022} & 41.7 & 41.7 & 66.6 & 76.8 & 12.3 \\
        & Hi-MPC\cite{rao_hierarchical_2024} & 47.5 & 47.5 & 70.3 & 78.6 & 17.4 \\
        & TransSG\cite{rao_transg_2023} & 68.7 & 68.7 & 86.5 & 91.8 & 30.1 \\
        & CMOT\cite{mukhtar_cmot_2024} & 77.8 & 77.8 & 91.2 & 97.3 & 77.6 \\
        & Ours & \textbf{100.0} & \textbf{96.4} & \textbf{100.0} & \textbf{100.0} & \textbf{97.4} \\
        \hline
        \end{tabular}}%
        \vspace{-0.5cm}
    \end{table}

    We also compare the performance of our proposed method with existing state-of-the-art techniques on the three datasets presented earlier. The results, summarized in Table \ref{tab:SOTA_comparaison}, demonstrate the superiority of our approach, particularly in leveraging RGB-D sequences over individual frames for person re-identification.
    
    \paragraph{\textbf{TVPR2 Dataset}} As shown in Table \ref{tab:SOTA_comparaison}, our method achieves superior results on the TVPR2 dataset compared to previous works, including methods that only use image-based inputs. The closest competitor, CMOT \cite{mukhtar_cmot_2024}, which also utilizes frame sequences, falls short of our performance. Our model achieves a Rank-1 accuracy of 99.5\%, which is further improved to 100.0\% when incorporating the Hungarian algorithm for more precise re-identification, demonstrating the robustness of our sequence-based approach.
    
    \paragraph{\textbf{GODPR Dataset}} On the GODPR dataset, our method outperforms the approach of Luna et al.\cite{luna_people_2021}, increasing the Rank-1 accuracy from 95.5\% to a perfect 100.0\%. Although the GODPR dataset has been used in fewer studies compared to others, it is particularly challenging due to the inclusion of infrared (IR) data alongside depth. This dataset highlights the flexibility of our approach in handling different modalities, and we will further explore the challenges posed by IR data in Section \ref{sec:ablation_study}.
    
    \paragraph{\textbf{BIWI RGBD-ID Dataset}} The BIWI RGBD-ID dataset is arguably the most challenging of the three due to its complex variations in appearance and environmental conditions. Nevertheless, our method significantly improves upon previous results, boosting the CMC-1 score from 77.8\% to 96.4\%, demonstrating the interest of sequence over single frame. Additionally, by applying the Hungarian algorithm, we achieve perfect re-identification, with a precision of 100.0\%, further demonstrating the strength of our approach in difficult re-identification scenarios.

\subsection{Ablation study}\label{sec:ablation_study}
\subsubsection{Impact of the Hungarian Algorithm on Precision}
    
    This ablation study investigates the effectiveness of the Hungarian algorithm in optimizing re-identification performance, particularly as the size of the distance matrix increases. We analyze this impact using the TVPR2 test subset with our depth-only model.

\begin{figure}
    \centering
    \includegraphics[width=0.4\textwidth]{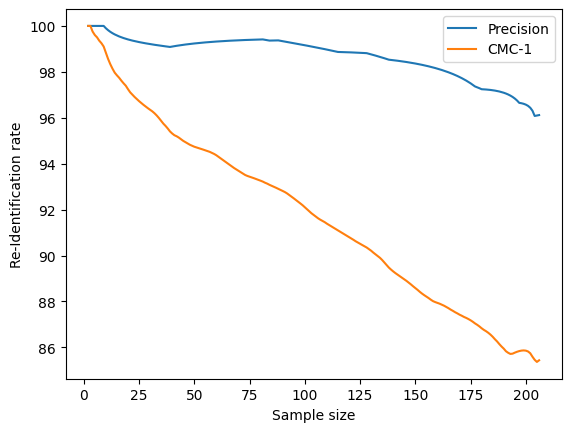}
    \caption{Impact of sample size on re-identification performance in the TVPR2 test subset. The plot shows how CMC-1 and precision decrease as the sample size increases. The precision metric benefits from the Hungarian algorithm, which maintains higher accuracy even in larger datasets.} \label{fig:reid_rate}
    \vspace{-0.5cm}
    \end{figure}

    As seen in Figure \ref{fig:reid_rate}, the CMC-1 metric, which reflects individual matching accuracy, shows a pronounced decline as the sample size increases, dropping from nearly 100\% for small test sets to around 86\% for larger sets. This decline highlights the challenges of maintaining high accuracy when evaluating matches at the individual level without accounting for global assignment optimization. Larger datasets inherently introduce more variability and ambiguity, such as individuals with similar appearances or noise, which complicate the re-identification process.

    In contrast, the precision metric remains more stable, with a smaller drop from near 100\% to approximately 96\%. Precision directly evaluates the effectiveness of the Hungarian algorithm \cite{kuhn_Hungarian_1955} in resolving the global assignment problem by minimizing the total cost in the distance matrix, as discussed in Section \ref{sec:assignment_problem}. This stability demonstrates the algorithm's ability to handle the increased complexity introduced by larger datasets, ensuring that matches are optimized holistically rather than relying on local, nearest-neighbor approaches.

    This study illustrates that the Hungarian algorithm is crucial for achieving robust re-identification performance in large-scale scenarios. By leveraging global optimization, it mitigates the limitations of CMC-1 and ensures consistent precision even as the sample size grows. These findings underscore the importance of the Hungarian algorithm in depth-only, privacy-preserving setups, where optimizing global assignments is critical for scalability and reliability.

\hfill\\[-0.7cm]
\subsubsection{Cross-Dataset Evaluation}
    
    To evaluate the generalization capability of our model, we performed cross-dataset experiments where the model trained on the TVPR2 dataset was directly tested on the GODPR and BIWI RGBD-ID datasets, without any additional fine-tuning. This experiment aimed to assess the robustness of our approach in more challenging, unseen scenarios. The results, displayed in Table \ref{tab:results_without_training}, highlight the performance of both the depth-only and RGB-D models in cross-dataset settings.
    
    $\bullet{}$ {\textbf{GODPR Dataset}} For the GODPR dataset, we observe an interesting result: the depth-only model outperforms the RGB-D model in terms of CMC-1 accuracy. Specifically, the depth-only model achieves a CMC-1 of 47.8\%, while the RGB-D model scores 43.3\%. This outcome is not entirely surprising, as the GODPR dataset contains IR frames rather than RGB data, which is incompatible with the RGB-D model trained on TVPR2's RGB frames. Even in this mismatched modality scenario, the performance remains reasonable. When applying the Hungarian algorithm, the CMC-1 improves further to 56.7\% for the depth-only model and 53.7\% for the RGB-D model, demonstrating the algorithm's ability to refine matches even when the model is not specifically trained for the dataset.
    
    $\bullet{}$ {\textbf{BIWI RGBD-ID Dataset}} For the BIWI RGBD-ID dataset, the results show the opposite trend. The RGB-D model significantly outperforms the depth-only model, achieving a CMC-1 of 67.9\% compared to 25.0\% for the depth model. This discrepancy can be attributed to the difference in camera viewpoints between the datasets: TVPR2 uses a top-view configuration, while BIWI RGBD-ID uses a frontal camera view. The depth model struggles to generalize from a top-down to a frontal view, which explains its lower performance. In contrast, the RGB-D model demonstrates better adaptability. The Hungarian algorithm further enhances the results, increasing the CMC-1 to 85.7\% for the RGB-D model and 50.0\% for the depth-only model, highlighting its effectiveness in difficult cross-dataset scenarios.
    
    These cross-dataset evaluations reveal the strengths and limitations of our models under mismatched data conditions and further underline the role of the Hungarian algorithm in improving performance, especially in challenging re-identification settings.
    
    \begin{table}
        \centering
        \caption{Results for GODPR and BIWI RGBD-ID datasets, trained only on TVPR2 dataset (in \%, higher is better)}\label{tab:results_without_training}
        \tabcolsep=0.11cm
        \resizebox{\columnwidth}{!}{%
        \begin{tabular}{|l|l|l|l|l|l|l|}
        \hline
        Dataset &  Modality & \textbf{Precision} & \textbf{Rank-1} & Rank-5 & Rank-10 & \textbf{mAP} \\
        \hline
        \hline
        GODPR & IR-D & 53.7 & 43.3 & 85.7 & 92.8 & 31.5 \\
        & Depth & 56.7 & 47.8 & 82.1 & 89.2 & 19.1 \\
        \hline
        BIWI RGBD-ID & RGB-D & 85.7 & 67.9 & 92.8 & 96.4 & 70.6 \\
        & Depth & 50.0 & 25.0 & 53.5 & 75.0 & 22.0 \\
        \hline
        \end{tabular}}
        \vspace{-0.5cm}
    \end{table}

\subsection{Computational Efficiency Discussion}

The proposed method demonstrates computational efficiency suitable for real-time applications. During inference, the model requires less than 3GB of GPU memory per person and takes under one second to extract embeddings from a sequence, outpacing real-time requirements.

Potential bottlenecks include the need for Regions of Interest (RoIs) as input, which depends on external detection and tracking systems outside the study's scope. Additionally, global re-identification is performed periodically, requiring a distance matrix and solving the assignment problem with the Hungarian algorithm, which has a time complexity of $O(n^3)$. While computationally intensive, this step is manageable due to its periodic nature.

This efficiency makes the approach practical for privacy-sensitive, real-time scenarios like public transport monitoring, balancing speed and accuracy effectively.

\section{Conclusion and Future Work}
\label{sec:conclusion}

In this paper, we proposed a novel approach for person re-identification using depth images with a strong focus on privacy preservation. By leveraging depth data, which inherently obscures facial features and other identifiable characteristics, our method offers a privacy-friendly alternative to traditional RGB-based Re-ID systems. The top-down configuration ensures that sensitive personal details are not captured, providing a valuable balance between privacy and behavioral data collection.
A central aspect of our approach is the use of the Hungarian algorithm to solve the association problem, optimizing matches by minimizing the global cost in the distance matrix. This method enhances re-identification performance, particularly in challenging scenarios such as occlusion or similar appearances. Additionally, we introduced the use of temporal sequences as inputs, which significantly improved the model’s ability to capture dynamic movement patterns, further boosting the accuracy of both depth-only and RGB-D models.
Our experimental results across multiple datasets demonstrated the superiority of our models over state-of-the-art methods, especially in privacy-sensitive applications. The combination of depth data with the Hungarian algorithm presents a compelling solution for practical applications, such as public transport systems, where privacy and accuracy are both paramount.

Future work includes improving real-time deployment, scaling to larger datasets, and integrating modalities like thermal or multi-camera data.

\section*{Acknowledgment}
This material is based upon work supported by the ANRT (Association nationale de la recherche et de la technologie) in France with a CIFRE fellowship granted to \href{http://www.explainconsultancy.com/en/}{Explain}.

\newpage
\section{Ethical Impact Statement}

In this paper, we propose a privacy-preserving approach for person re-identification using depth images captured from a top-view perspective. The top-down depth imaging method inherently obscures identifiable features, such as faces, ensuring that individuals cannot be easily recognized based on personal or facial information. This feature makes our method particularly suitable for deployment in sensitive settings, such as public transport hubs, where privacy is paramount. By using depth images instead of traditional RGB data, we further mitigate privacy concerns typically associated with surveillance, as depth data is less prone to revealing detailed identity characteristics.

Our methodology addresses person re-identification through the Hungarian algorithm to match individuals entering and exiting a monitored area. This matching process does not involve storing personal data but focuses on tracking movement patterns, which are valuable for improving crowd management and urban infrastructure without compromising individual privacy. Additionally, our reliance on batch hard triplet loss and transformer-based sequence encoding enhances re-identification accuracy without increasing the risk of exposing personal information.

While the proposed method is designed to preserve privacy, concerns may arise regarding data consent, bias, and the broader implications of surveillance. To address these issues:
\begin{itemize}
\item Data Consent: Researchers must ensure that data collection is conducted transparently, with explicit consent from individuals whenever possible, even when using depth images that lack identifiable features.
\item Bias: The reliance on specific datasets, such as those with limited diversity in demographics or movement patterns, may introduce biases into the model. Future work should prioritize diverse and representative datasets to mitigate these biases and ensure equitable performance across populations.
\item Surveillance Implications: Although the method does not capture identifiable features, its deployment in surveillance contexts must be carefully monitored to prevent misuse, such as unauthorized tracking or profiling. Policies should be established to limit the scope and duration of data storage and use.
\end{itemize}

\textbf{Ethical Review Approval}
This study did not undergo formal ethical review board approval because it does not involve direct interaction with human subjects or the collection of personally identifiable information. The data collected consists exclusively of depth images from a top-view perspective, which inherently lacks facial or other identifiable features, thereby minimizing privacy risks. Since the research relies solely on anonymized depth data and follows a privacy-preserving approach without the recording of sensitive information, it is not subject to the requirements for oversight by a formal ethical review board.

{\small
\bibliographystyle{ieee}
\bibliography{biblio}
}

\end{document}